\documentclass{article}

\usepackage{arxiv}

\usepackage[utf8]{inputenc} 
\usepackage[T1]{fontenc}    
\usepackage{hyperref}       
\usepackage{url}            
\usepackage{booktabs}       
\usepackage{amsfonts}       
\usepackage{nicefrac}       
\usepackage{microtype}      
\usepackage{lipsum}		
\usepackage{graphicx}
\usepackage{natbib}
\usepackage{doi}

\usepackage{listings}
\usepackage{xcolor}

\definecolor{codegreen}{rgb}{0,0.6,0}
\definecolor{codegray}{rgb}{0.5,0.5,0.5}
\definecolor{codepurple}{rgb}{0.58,0,0.82}
\definecolor{backcolour}{rgb}{0.95,0.95,0.92}
\definecolor{elenscolor}{HTML}{029E73}
\definecolor{folcolor}{HTML}{CFBF00}

\lstdefinestyle{mystyle}{
    backgroundcolor=\color{backcolour},   
    commentstyle=\color{codegreen},
    keywordstyle=\color{magenta},
    numberstyle=\tiny\color{codegray},
    stringstyle=\color{codepurple},
    basicstyle=\ttfamily\footnotesize,
    breakatwhitespace=false,         
    breaklines=true,                 
    captionpos=b,                    
    keepspaces=true,                 
    numbers=left,                    
    numbersep=5pt,                  
    showspaces=false,                
    showstringspaces=false,
    showtabs=false,                  
    tabsize=2
}
\hypersetup{colorlinks=true}

\title{\emph{PyTorch, Explain!}\\A Python Library for Logic Explained Networks}

\author{ \href{https://orcid.org/0000-0003-3155-2564}{\includegraphics[scale=0.06]{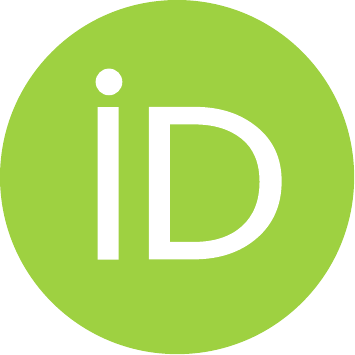}\hspace{1mm}Pietro Barbiero} \\
	Department of Computer Science\\
	University of Cambridge\\
	Cambridge, UK \\
	\texttt{pb737@cam.ac.uk} \\
	\And
	\href{https://orcid.org/0000-0003-3155-2564}{\includegraphics[scale=0.06]{orcid.pdf}\hspace{1mm}Gabriele Ciravegna} \\
	Department of Computer Science\\
	Universita' di Siena\\
	Siena, IT \\
	\texttt{gabriele.ciravegna@unifi.it} \\
	\And
	Dobrik Georgiev \\
	Department of Computer Science\\
	University of Cambridge\\
	Cambridge, UK \\
	\texttt{dgg30@cam.ac.uk} \\
	\And
	Franscesco Giannini \\
	Department of Computer Science\\
	Universita' di Siena\\
	Siena, IT \\
	\texttt{fgiannini@diism.unisi.it} \\
}

\renewcommand{\shorttitle}{\textit{arXiv} Template}

\hypersetup{
pdftitle={A template for the arxiv style},
pdfsubject={q-bio.NC, q-bio.QM},
pdfauthor={David S.~Hippocampus, Elias D.~Striatum},
pdfkeywords={First keyword, Second keyword, More},
}

\begin{document}
\maketitle

\begin{abstract}
\textit{PyTorch, Explain!} is a Python module integrating a variety of state-of-the-art approaches to provide logic explanations from neural networks. This package focuses on bringing these methods to non-specialists. It has minimal dependencies and it is distributed under the Apache 2.0 licence allowing both academic and commercial use. Source code and documentation can be downloaded from the github repository: \url{https://github.com/pietrobarbiero/pytorch_explain}.
\end{abstract}


\section{Introduction}
The lack of transparency in the decision process of some machine learning models, such as neural networks, limits their application in many safety-critical domains \citep{chander2018working}. Employing black-box\footnote{In the context of this paper, a black-box classifier is any classifier that cannot provide human understandable explanations about its decisions} models may be unacceptable in contexts such as industry, medicine or courts, where the potential economical or ethical repercussions are calling for lawmakers to discourage from a reckless application of non-interpretable models \citep{gdpr2017,law10code,goddard2017eu,gunning2017explainable}. As a consequence, research in Explainable Artificial Intelligence (XAI) has become strategic and has been massively encouraged, leading to the development of a variety of techniques that aim at explaining black-box models \citep{das2020opportunities,brundage2020toward} or at developing effective interpretable models \citep{carvalho2019machine,rudin2019stop}.

The need for high-quality human-friendly explanations is one of the main reasons why concept-based explanations are receiving ever-growing consideration. Explanations are given in terms of human-understandable symbols (the \textit{concepts}) rather than raw features such as pixels or characters \citep{kim2018tcav,ghorbani2019towards,koh2020concept}. As a consequence, they seem more suitable to serve many strategic human purposes such as decision making tasks. For instance, a concept-based explanation may describe a high-level category through its attributes as in ``a \textit{human} has \textit{hands} and a \textit{head}''.
While collecting high-quality evidences for an explanation is a common feature of concept-based techniques, there are very few approaches formulating hypothesis and even less providing synthetic descriptions whose validity can be quantitatively assessed \citep{das2020opportunities}.
A possible solution is to rely on a formal language that is
very expressive and closely related to natural language and reasoning, such as First-Order Logic (FOL). A FOL explanation can be considered a special kind of a concept-based explanation, where the description is given in terms of logical predicates, connectives and quantifiers, such as ``$\forall x:\ \textit{is\_human}(x) \rightarrow \textit{has\_hands}(x) \wedge \textit{has\_head}(x)$''. However, FOL formulas generally express much more complex relationships among the concepts 
involved in a certain explanation. Compared to other concept-based techniques, logic-based explanations provide many key advantages, that we briefly described in what follows.
An explanation reported in FOL is a rigorous and unambiguous statement (\textsc{clarity}). This formal clarity may serve cognitive-behavioral purposes such as engendering trust, aiding bias identification, or taking actions/decisions. For instance, dropping quantifiers and variables for simplicity, the formula ``\textit{snow} $\wedge$ \textit{tree} $\leftrightarrow$ \textit{wolf}'' may easily outline the presence of a bias in the collection of training data.
Different logic-based explanations can be combined to describe groups of observations or global phenomena (\textsc{modularity}). 
For instance, for an image showing only the face of a person, an explanation could be ``(\textit{nose} $\wedge$ \textit{lips}) $\rightarrow$ \textit{human}'', while for another image showing a person from behind a valid explanation could be ``(\textit{feet} $\wedge$ \textit{hair} $\wedge$ \textit{ears})  $\rightarrow$ \textit{human}''. The two local explanations can be combined into ``(\textit{nose} $\wedge$ \textit{lips}) $\vee$ (\textit{feet} $\wedge$ \textit{hair} $\wedge$ \textit{ears}) $\rightarrow$ \textit{human}''.
The quality of logic-based explanations can be quantitatively measured
to check their correctness and completeness (\textsc{measurability}). For instance, once the explanation ``(\textit{nose} $\wedge$ \textit{lips}) $\vee$ (\textit{feet} $\wedge$ \textit{hair} $\wedge$ \textit{ears})'' is extracted for the class \textit{human}, this logic formula can be applied on a test set to check its generality in terms of quantitative metrics like accuracy, fidelity and consistency.
Finally, FOL-based explanations can be rewritten in
    different equivalent forms such as in \emph{Disjunctive Normal Form} (DNF) and \emph{Conjunctive Normal Form} (CNF)  (\textsc{simplifiability}). Further, techniques such as the Quine–McCluskey algorithm can be used to compact and simplify logic explanations \citep{mccoll1878calculus,quine1952problem,mccluskey1956minimization}. For instance, the explanation ``(\textit{person} $\wedge$ \textit{nose}) $\vee$ ($\neg$\textit{person} $\wedge$ \textit{nose})'' can be easily simplified in ``\textit{nose}''.

This work presents a Python library for XAI enabling neural networks to \textit{solve and explain} a categorical learning problem integrating elements from deep learning and logic.
Differently from vanilla neural architectures, these models can be directly interpreted by means of a set of FOL formulas. In order to implement such a property, such models require their inputs to represent the activation scores of human-understandable concepts. Then, specifically designed learning objectives allow them to make predictions in a way that is well suited for providing FOL-based explanations that involve the input concepts. In order to reach this goal, LENs exploit parsimony criteria aimed at keeping their structure simple as described in recent works \citep{koh2020concept,ciravegna2020constraint,ciravegna2020human}.

\section{Background}
Classification is the problem of identifying a set of categories $y \in Y\subset[0,1]^r$ an observation $x \in X\subset\mathbb{R}^d$ belongs to. A standard neural network is a black-box model $f: X \mapsto Y$ predicting for any sample $x\in X$ the corresponding class membership $\hat{y}\in Y$. In case the input features $x$ are not easily interpretable (low-level feature, image pixels) concept-based classifiers have been introduced to predict class memberships $Y$ from human-understandable categories  (a.k.a. \textit{concepts}) $C\subset[0,1]^k$: $f: C \mapsto Y$ to improve the understanding of black boxes and their decision process. Concepts can either correspond to the predictions of a classifier (i.e. $g: X \mapsto C$) \citep{koh2020concept} or simply to a re-scaling of the inputs space from the unbounded $\mathbb{R}^d$ to the unit interval $[0, 1]^k$ such that input features can be treated as logic predicates. Concept-based classifiers improve human understanding as their input and output spaces consists of interpretable symbols. 

Recent work on concept-based neural networks has led to the development of models like the $\psi$ network \citep{ciravegna2020human} or the entropy-based network \cite{barbiero2021entropy} i.e., concept-based classifiers \textit{explaining their own decision process}. These models are designed to compute concise logic formulas representing \emph{how} the network combines input concepts in order to arrive to a prediction. 

This library implements the core learning criteria and methods allowing a customized implementation of neural models providing logic explanations.

\section{Application Programming Interfaces (APIs)}
The code library is designed with intuitive APIs requiring only a few lines of code to train and get explanations from deep neural networks. The library currently provides the APIs for three explainable neural architectures: $\psi$ networks described in \citet{ciravegna2020constraint}, multi-layer neural networks with rectified linear units, and entropy-based networks \cite{barbiero2021entropy}.

The library supports a fine-grained customization of the neural networks as shown in the following code example \ref{code:example2}. The architecture of the model and the training loop is defined by means of standard PyTorch APIs  \citep{paszke2019pytorch}. The \texttt{entropy\_logic\_loss} allows the network to get rid of less relevant input concepts and to generate concise explanations for each prediction \citep{ciravegna2020constraint}. The \texttt{explain\_class} method allows the extraction of logic formulas from the trained model. Once extracted, formulas can be tested on an unseen set of test samples using the \texttt{test\_explanation} method providing the accuracy of the logic formula.
Explanations will be logic formulas in disjunctive normal form. In the example below, the explanation will be $y=1 \leftrightarrow (x_1 \wedge \neg x_2) \vee (x_2  \wedge \neg x_1)$ corresponding to $y=1 \leftrightarrow x_1 \oplus x_2$.

\begin{lstlisting}[language=Python, label=code:example2, caption=Example on how to use \emph{PyTorch, Explain!} APIs.]
import torch
from torch.nn.functional import one_hot
import torch_explain as te
from torch_explain.nn.functional import entropy_logic_loss
from torch_explain.logic.nn import entropy
from torch_explain.logic.metrics import test_explanation, complexity

# train data
x_train = torch.tensor([
    [0, 0],
    [0, 1],
    [1, 0],
    [1, 1],
], dtype=torch.float)
y_train = torch.tensor([0, 1, 1, 0], dtype=torch.float).unsqueeze(1)

# instantiate an "entropy-based network"
layers = [
    te.nn.EntropyLinear(x_train.shape[1], 10, n_classes=2),
    torch.nn.LeakyReLU(),
    torch.nn.Linear(10, 4),
    torch.nn.LeakyReLU(),
    torch.nn.Linear(4, 1),
]
model = torch.nn.Sequential(*layers)

# fit the model
optimizer = torch.optim.AdamW(model.parameters(), lr=0.01)
loss_form = torch.nn.CrossEntropyLoss()
model.train()
for epoch in range(1001):
    optimizer.zero_grad()
    y_pred = model(x_train).squeeze(-1)
    loss = loss_form(y_pred, y_train) + 0.00001 * entropy_logic_loss(model)
    loss.backward()
    optimizer.step()

# get first-order logic explanations for a specific
y1h = one_hot(y_train.squeeze().long())
explanation = entropy.explain_class(model, x_train)

# compute explanation accuracy and complexity
accuracy, preds = test_explanation(explanation, x_train, y1h, target_class=1)
explanation_complexity = complexity(explanation)
\end{lstlisting}








\section{Software and documentation availability}
In order to make state-of-the-art approaches accessible to the whole community, we released this library as a Python package on PyPI: \url{https://pypi.org/project/torch-explain/}. An extensive documentation on methods is available on read the docs\footnote{\url{https://pytorch-explain.readthedocs.io/en/latest/}.} and unit tests results on TravisCI\footnote{\url{https://travis-ci.org/pietrobarbiero/pytorch_explain/}.}. The Python code and the scripts used for benchmarking against state-of-the-art white-box models, including parameter values and documentation, is freely available under Apache 2.0 Public License from a GitHub repository\footnote{\url{https://github.com/pietrobarbiero/logic\_explainer\_networks}.}.

\section{Conclusion}

\subsection{Limitations and future research directions}
The extraction of a first-order logic explanation requires symbolic input and output spaces. This constraint is the main limitation of our framework, as narrows the range of applications down to symbolic I/O problems. In some contexts, such as computer vision, the use of LENs may require additional annotations and attribute labels to get a consistent symbolic layer of concepts. However, recent work on automatic concept extraction may partially solve this issue leading to more cost-effective concept annotations \citep{ghorbani2019towards,kazhdan2020now}.

The improvement of LENs models is an open research area. The efficiency and the classification performances of fully interpretable LENs, i.e. $\psi$ network, is still quite limited due to the extreme pruning strategy.

\subsection{Broader impact}
Current legislation in US and Europe binds AI to provide explanations especially when the economical, ethical, or financial impact is significant \citep{gdpr2017,law10code}. This work contributes to a lawful and safer adoption of some of the most powerful AI technologies allowing deep neural networks to have a greater impact on society. Extracting first-order logic explanations from deep neural networks enables satisficing \citep{simon1956rational} knowledge distillation while achieving performances comparable with the state of the art. The formal language of logic provides clear and synthetic explanations, suitable for laypeople, managers, and in general for decision makers outside the AI research field.

Thanks to their explainable nature, LENs can be effectively used to understand the behavior of an existing algorithm, to reverse engineer products, to find vulnerabilities, or to improve system design. From a scientific perspective, formal knowledge distillation from state-of-the-art networks may enable scientific discoveries or confirmation of existing theories.

\section*{Acknowledgments and Disclosure of Funding}
We thank Stefano Melacci, Pietro Li\'o, and Marco Gori for useful feedback and suggestions.

This work was partially supported by the European Union’s Horizon 2020 research and innovation programme under grant agreement No 848077.

\bibliographystyle{apalike}
\bibliography{references}  






\end{document}